\documentclass{article}

\usepackage{arxiv}

\usepackage[utf8]{inputenc} 
\usepackage[T1]{fontenc}    
\usepackage{hyperref}       
\usepackage{url}            
\usepackage{booktabs}       
\usepackage{amsfonts}       
\usepackage{nicefrac}       
\usepackage{microtype}      
\usepackage{lipsum}		
\usepackage{graphicx}
\usepackage{natbib}
\usepackage{doi}
\usepackage{multirow}
\usepackage{authblk}

\author[1,*]{Jihoon Kweon}
\author[2,*]{Kyunghwan Kim}
\author[2]{Chaehyuk Lee}
\author[2]{Hwi Kwon}
\author[2]{Jinwoo Park}
\author[2]{Kyoseok Song}
\author[3]{Young In Kim}
\author[3]{Jeeone Park}
\author[4]{Inwook Back}
\author[5]{Jae-Hyung Roh}
\author[1]{Youngjin Moon}
\author[6]{Jaesoon Choi}
\author[4,$\dagger$]{Young-Hak Kim}
\affil[1]{Departments of Convergence Medicine, Asan Medical Center, University of Ulsan College of Medicine, Seoul, Republic of Korea}
\affil[2]{Medipixel, Inc., Seoul, Republic of Korea}
\affil[3]{Department of Medical Science, Asan Medical Institute of Convergence Science and Technology, Asan Medical Center, University of Ulsan College of Medicine, Seoul, Republic of Korea}
\affil[4]{Division of Cardiology, Department of Internal Medicine, Asan Medical Center, University of Ulsan College of Medicine, Seoul, Republic of Korea}
\affil[5]{Department of Cardiology, Chungnam National University Sejong Hospital, Chungnam National University School of Medicine, Daejeon, Republic of Korea}
\affil[6]{Department of Biomedical Engineering, Asan Medical Center, University of Ulsan College of Medicine, Seoul, Republic of Korea}
\affil[ ]{ }
\affil[*]{These authors contributed equally as first author}
\affil[$\dagger$]{Correspondence: mdyhkim@amc.seoul.kr}
\title{Deep reinforcement learning for guidewire navigation in coronary artery phantom}

\renewcommand{\headeright}{ }
\renewcommand{\undertitle}{ }
\renewcommand{\shorttitle}{Deep reinforcement learning for guidewire navigation in coronary artery phantom}

\hypersetup{
pdftitle={Deep RL for guidewire navigation},
pdfsubject={Reinforcement learning},
pdfauthor={J. Kweon, K. Kim, et al.},
pdfkeywords={Reinforcement learning; Coronary intervention; Guidewire navigation},
}

\begin{document}
\maketitle

\begin{abstract}
In percutaneous intervention for treatment of coronary plaques, guidewire navigation is a primary procedure for stent delivery. Steering a flexible guidewire within coronary arteries requires considerable training, and the non-linearity between the control operation and the movement of the guidewire makes precise manipulation difficult. Here, we introduce a deep reinforcement learning (RL) framework for autonomous guidewire navigation in a robot-assisted coronary intervention. Using Rainbow, a segment-wise learning approach is applied to determine how best to accelerate training using human demonstrations with deep Q-learning from demonstrations (DQfD), transfer learning, and weight initialization. ‘State’ for RL is customized as a focus window near the guidewire tip, and subgoals are placed to mitigate a sparse reward problem. The RL agent improves performance, eventually enabling the guidewire to reach all valid targets in ‘stable’ phase. Our framework opens a new direction in the automation of robot-assisted intervention, providing guidance on RL in physical spaces involving mechanical fatigue.
\end{abstract}


\section{Introduction}
\label{sec:intro}
\begin{figure}
	\centering
    \includegraphics[width=\textwidth,keepaspectratio]{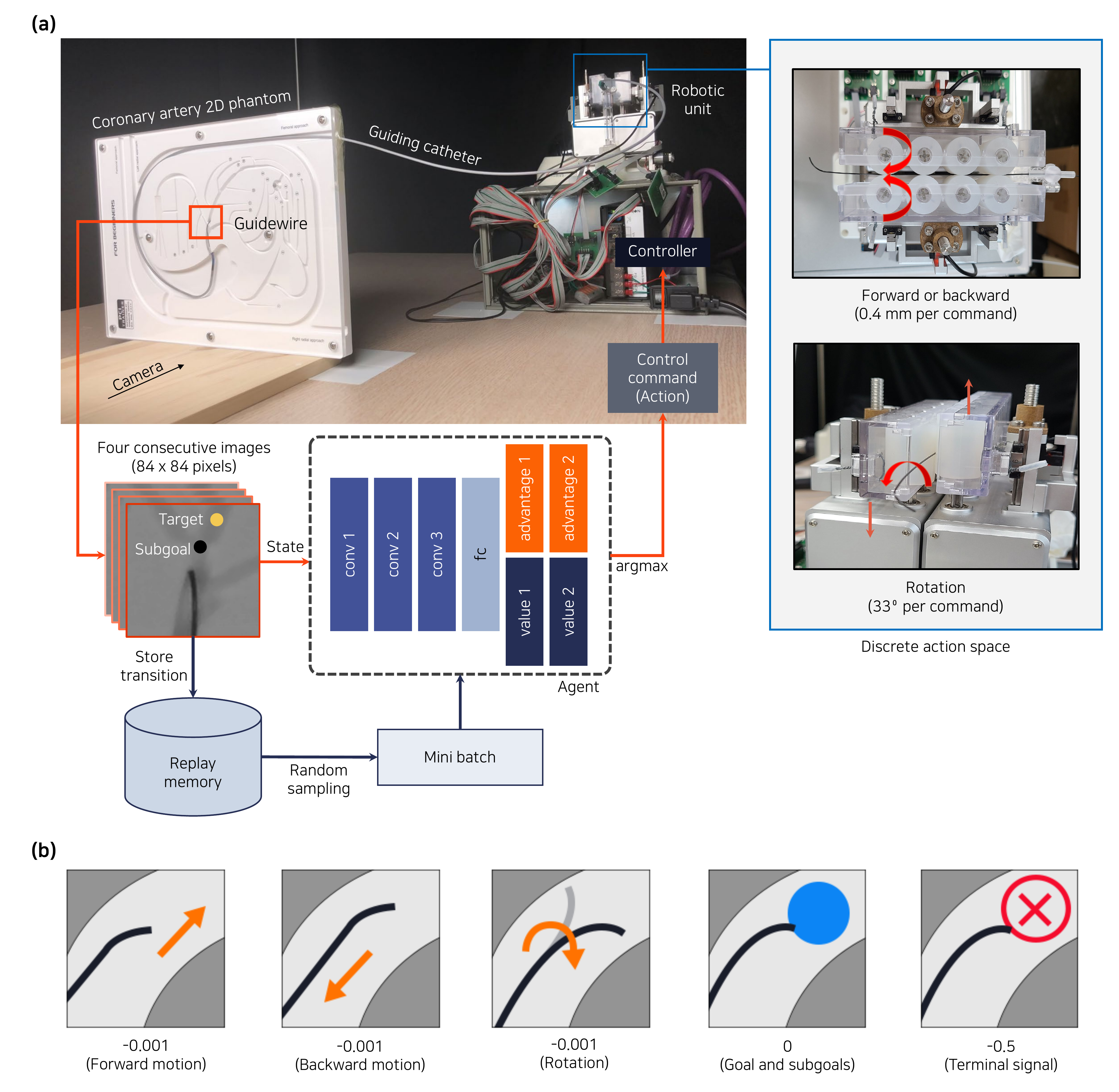}
	\caption{(a) Interaction between physical environment and RL agent. Of the entire navigation area, only information around the guidewire tip is defined as ‘state’, which is given as an input to the RL agent. The RL agent selects a control command as an ‘action’ maximizing the expected ‘rewards’ at the current 'state', and the selected control command is transferred to the robotic device to perform one of the control operations: forward/backward motion and rotation. While this process is repeated, sets of states, actions, rewards, and next states (transitions) are accumulated in the replay memory, and RL training is performed periodically using transitions. (b) Reward design of reinforcement learning for guidewire navigation.}
	\label{fig:fig1}
\end{figure}

Coronary arteries are vessels that supply oxygen-rich blood and nutrients to the myocardium. When coronary arteries are obstructed, the heart muscle is not supplied with sufficient oxygen and energy, resulting in ischemia. Such ischemic heart disease is reported to be the leading cause of death, responsible for 16$\%$ of the world’s total deaths.\\
Percutaneous coronary intervention (PCI) with balloon angioplasty and stent implantation is the standard treatment for coronary artery stenosis. The catheter access provides a path from the incision area to the coronary ostium, and the balloon and stent are delivered along the guidewire to the target location. Because the diameter of the coronary artery is relatively small ($\leq$4 mm) (\citealp{dodge1992lumen}) and the distance between the operating controls and the distal end of the guidewire is long, a considerable amount of specialized training is required for precise manipulation of the interventional devices. Friction with the vessel wall causes deformation of the flexible tip, impeding guidewire control, and a risk of perforation may increase in the treatment procedure of severe lesions (\citealp{guttmann2017prevalence}). The non-linear relationship between the control motion applied to the guidewire and the movement of the distal end is an important feature that makes the device difficult to precisely control.\\
Interventional robots for coronary diseases have been introduced for improved manipulation of interventional devices with reduced irradiation (\citealp{beyar2005concept}; \citealp{kiemeneij2008use}). The safety and feasibility of interventional robots have been demonstrated by clinical studies (\citealp{weisz2013safety}; \citealp{patel2020comparison}), and their applications have been widened to complex lesions such as multi-vessel diseases and chronic total occlusion (\citealp{hirai2020initial}; \citealp{mahmud2017demonstration}). Integration of telecommunication systems with robotic apparatus enables the remote operation of robot-assisted PCI (\citealp{patel2019long}). In pandemic situations such as current COVID-19 wave, robotic procedures have been proposed as a way to reduce the potential infection risk for medical staff and patients (\citealp{attanasio2021autonomy}; \citealp{zemmar2020rise}). With the adoption of artificial intelligence, interventional robots are expected to be further automated to minimize interference from human operators (\citealp{sardar2019impact}).\\
Reinforcement learning (RL) is an area of machine learning that trains an agent to achieve a goal by maximizing rewards, which are imposed by the next state in response to changing conditions when an action is taken in the current state. Deep RL has been applied to various domains, such as Go and computer games, to surpass the world's best human players (\citealp{silver2016mastering}; \citealp{mnih2015human}), and its applications have expanded from software to the control engine for real-world hardware (\citealp{levine2016end}; \citealp{gandhi2017learning}; \citealp{pinto2017asymmetric}). Considering that simple control operations are repeatedly performed to utilize a robot-assisted intervention system, deep RL may be a solution that effectively alleviates the burden of human operators. Recent applications for autonomous control of interventional devices in phantom simulation supported the potential applicability of deep RL (\citealp{you2019automatic}; \citealp{behr2019deep}; \citealp{karstensen2020autonomous}; \citealp{chi2020collaborative}; \citealp{zhao2019cnn}).\\
In this study, we propose a deep RL framework for autonomous guidewire navigation in robot-assisted coronary interventions. We focus on how to accelerate the RL training to prevent mechanical fatigue on the guidewire due to repetitive movements. First, under the constraints of the discrete action space and training in a real-world setting, Rainbow (\citealp{hessel2018rainbow}) was applied (Figure \ref{fig:fig1}a). Rainbow integrating Deep Q-Networks (DQN) (\citealp{mnih2015human}) with recent advancements (\citealp{van2016deep}; \citealp{schaul2015prioritized}; \citealp{wang2016dueling}; \citealp{fortunato2017noisy}; \citealp{bellemare2017distributional}) in reinforcement learning has demonstrated outstanding performance in real world environments (\citealp{church2020deep}). Replay memory (\citealp{mnih2015human}), a representative an off-policy method to enhance sample efficiency and increase the training speed with quality data, was a key component to reduce the physical time requirement. To optimize the initial composition of the replay memory, human demonstrations with DQfD (\citealp{hester2018deep}) and weighted random action (WRA) were evaluated. Second, the state for the RL agent was customized with a focus window near the guidewire tip. Given that the movement of the guidewire per control command was about two orders of magnitude smaller than the travel distance of the navigation, the focus window allowed the RL agent to confine its input to the area with the most important information. Subgoals like the dots in 'Pac-Man' guided the navigation to the goal beyond the focus window as well as mitigating the sparse reward problem. Finally, segment-wise training was conducted, which was inspired by the concept of curriculum learning (\citealp{graves2017automated}. When expanding the navigation area, transfer learning was applied using the model from previous training. Our framework was assessed in a two-dimensional (2D) coronary phantom for trainees and further validated in three-dimensional (3D) coronary phantom with fluid flow.

\section{Materials and Methods} \label{sec:materials_methods}
\subsection{Physical environment for reinforcement learning} \label{subsec:physical_env}
A robotic module developed in Asan Medical Center was used for guidewire navigation (Figure \ref{fig:fig1}a). A pair of roller units rotating in the opposite direction enabled forward and backward motion of the guidewire (Terumo Radifocus 0.035”, Terumo Co., Ltd., Japan). The vertical translation of the roller units produced the rotation of the guidewire. The roller units driven by step motors generated discrete control commands corresponding to 0.4 mm displacement or 33° rotation of the guidewire at the roller side. The guidewire with a pre-angled tip was delivered via the guiding catheter (Heartrail II JL-3.5, Terumo Co., Ltd., Tokyo, Japan) engaged at the ostium of the coronary artery in the phantom. The entire phantom area was captured by a RealSense™ D435 camera (Intel Co., Ltd., CA) mounted orthogonal to the phantom.

\subsection{Network setting for reinforcement learning}
\label{subsec:network_setting}

\subsubsection{Training}
\label{subsubsec:train}
To build an RL agent determining control commands using Rainbow (\citealp{hessel2018rainbow}), a convolutional neural network (CNN) was constructed (Figure \ref{fig:fig1}a). Using ‘state’ information composed of four consecutive images as input, the trained network output a distribution of Q values for deciding an ‘action’. According to the control command, the guidewire was manipulated by the robotic module and the RL agent receives a 'reward' depending on the 'next state'. A step was defined as the generation process of a transition, which was a set of state, next state, action, and reward. Every transition was saved in the replay memory.\\
In each episode, the guidewire tip initially located in front of the catheter was moved toward a goal by the combination of control commands. When the guidewire reached a target location within 500 steps, the episode was considered a ‘success’; otherwise it was considered a ‘failure’. After finishing an episode, the guidewire was pulled back to the initial location, and then a new episode began. The training consisted of 1000 episodes, and the goal was set to switch randomly for each episode. At the beginning of training, transitions for the replay memory were generated using weighted random action (WRA) or transferred weight for a given number of steps. The network was not updated for this ‘transition generation’ phase. The composition and generation method of transitions for the replay memory is summarized in Table \ref{tab:table1}.\\
The loss function was defined as a combination of Rainbow loss (\citealp{hessel2018rainbow}) with large margin classification loss and L2 regularization loss from DQfD (\citealp{hester2018deep}). The large margin classification loss was only used for training the RL agent with human demonstrations. Hyper-parameters used for the training are summarized in Table \ref{tab:atable1} in Appendix.

\begin{figure}
	\centering
    \includegraphics[width=\textwidth,keepaspectratio]{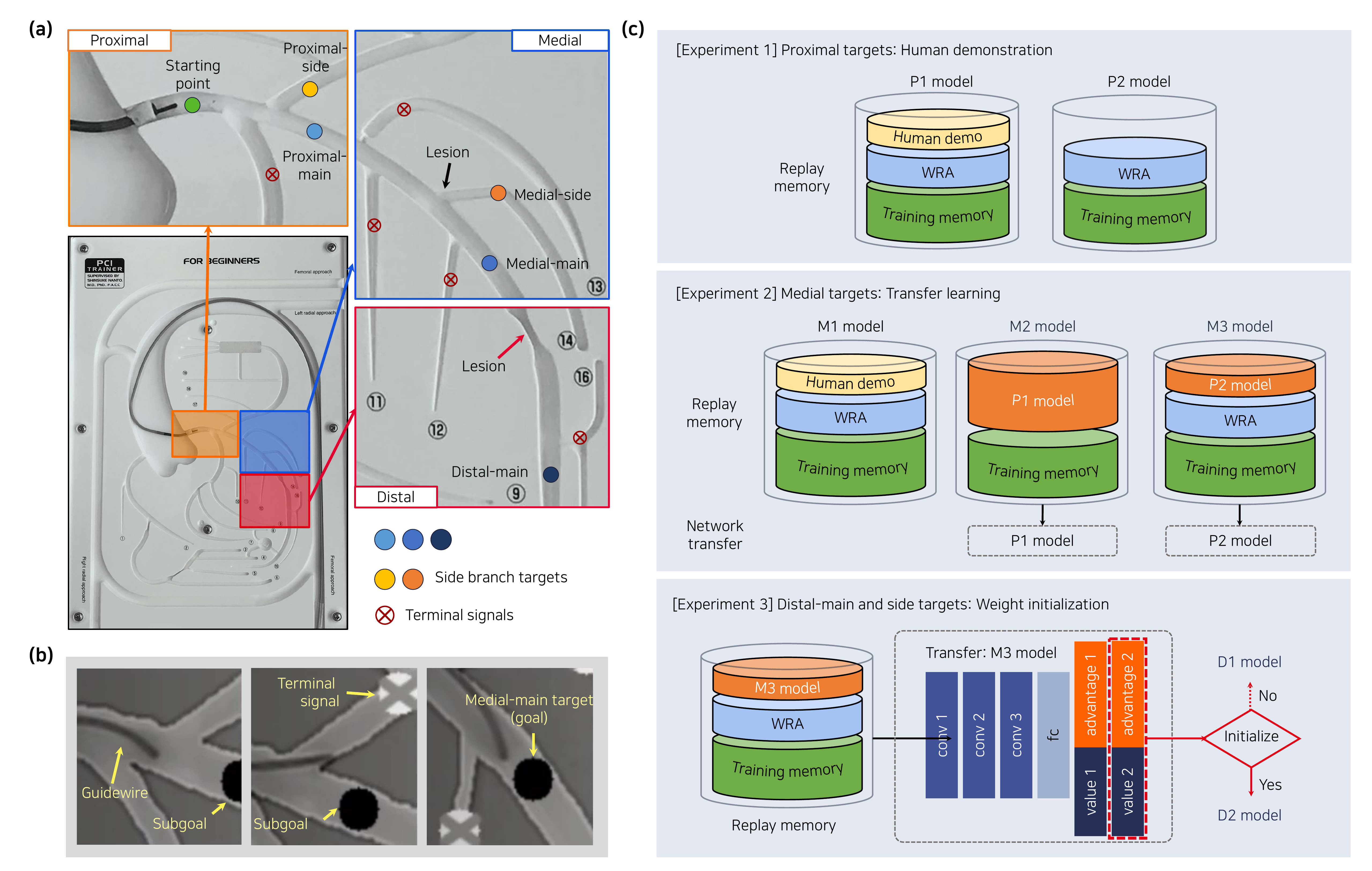}
	\caption{(a) The navigation area, divided into three zones, is initially designated as a proximal zone and is expanded by adding medial and distal zones, respectively. The goal is set at the target location of the guidewire, and terminal signals are assigned to other branches. (b) Because the goal is not visible in the focus window around the guidewire tip, subgoals are introduced. The focus window contains at least one subgoal or a goal. (c) Experimental setup according to navigation area.}
	\label{fig:fig2}
\end{figure}

\begin{table}
	\caption{Composition of initial replay memory. For P1 and M1 models, 3234 transitions from 40 episodes of human demonstrations are included in the replay memory prior to weighted random action (WRA). The probability of forward command in WRA is increased in experiment 3, because the distal zone has only a small branch that the guidewire is very unlikely to enter. Prox, proximal; Med, medial; Dist, distal; HD, human demonstration with DQfD; TR, transfer learning; WI, weight initialization.}
	\centering
	\begin{tabular}{cccccccc}
		\toprule
		{ }&{ }&{ }&{ }&\multicolumn{4}{c}{Generation of initial transitions in replay memory}                   \\
		\cmidrule(r){5-8}
		
		{Exp.} &
		{Target} &
		{Model} &
		{Feature} &
		{\begin{tabular}{@{}c@{}}Number of\\transitions\end{tabular}} &
		{\begin{tabular}{@{}c@{}}Generation\\method\end{tabular}} &
		{Segment} &
		{\begin{tabular}{@{}c@{}}Probability of\\command in WRA\end{tabular}}\\

		\midrule
		\multirow{2}*{1} &
		\multirow{2}*{\begin{tabular}{@{}c@{}}Prox-main,\\Prox-side\end{tabular}} &
		{P1} &
		{HD} &
		{10,000} &
		{WRA} &
		{Prox} &
		{0.4, 0.2, 0.4}\\
		\cmidrule(r){3-8}		
		&&
		{P2} &
		{-} &
		{10,000} &
		{WRA} &
		{Prox} &
		{0.4, 0.2, 0.4}\\
		
        \midrule

		\multirow{4}*{2} &
		\multirow{4}*{\begin{tabular}{@{}c@{}}Med-main,\\Med-side\end{tabular}} &
		{M1} &
		{HD} &
		{10,000} &
		{WRA} &
		{Prox, Med} &
		{0.4, 0.2, 0.4}\\
		
		\cmidrule(r){3-8}
		&&
		{M2} &
		{TR} &
		{10,000} &
		{P1} &
		{Prox, Med} &
		{-}\\
		
		\cmidrule(r){3-8}
		&&
		\multirow{2}*{M3} &
		\multirow{2}*{TR} &
		\multirow{2}*{10,000} &
		{P2} &
		{Prox} &
		{-}\\
		
		\cmidrule(r){6-8}		
		&&&&&
		{WRA} &
		{Med} &
		{0.4, 0.2, 0.4}\\

        \midrule

		\multirow{4}*{3} &
		\multirow{4}*{\begin{tabular}{@{}c@{}}Dist-main,\\Med-side,\\Prox-side\end{tabular}} &
		\multirow{2}*{D1} &
		\multirow{2}*{TR} &
		\multirow{2}*{20,000} &
		{M3} &
		{Prox, Med} &
		{-}\\

		\cmidrule(r){6-8}
		&&&&&
		{WRA} &
		{Dist} &
		{0.6, 0.2, 0.2}\\

		\cmidrule(r){3-8}
		&&
		\multirow{2}*{D2} &
		\multirow{2}*{TR, WI} &
		\multirow{2}*{20,000} &
		{M3} &
		{Prox, Med} &
		{-}\\
		\cmidrule(r){6-8}
		
		&&&&&
		{WRA} &
		{Dist} &
		{0.6, 0.2, 0.2}\\

		\bottomrule
	\end{tabular}
	\label{tab:table1}
\end{table}

\subsubsection{State}
\label{subsubsec:state}
In defining ‘state’, two major modifications of focus window and subgoal were introduced. The image area for the state, which was converted to grayscale as in X-ray angiography, was cropped to 84 × 84 pixels near the guidewire tip, allowing the RL agent to focus on more important information (Figure \ref{fig:fig2}b). The main drawback of the image crop was that the RL agent could not recognize the goal until the guidewire approached the target location. Therefore, subgoals were added on the path leading to the target location. The subgoals were initially set at the bifurcation points and additional subgoals were placed at a distance of 20 pixels, which is about a quarter of the image size for the state. Also, terminal signals were designated near the entrance of untargeted branches, which helped prevent the RL agent from making useless exploration.

\subsubsection{Action}
\label{subsubsec:action}
The action space of the RL agent was composed of forward/backward motion and rotation. The magnitude of each action generated by the robotic manipulator was fixed as constant. The rotational direction of the guidewire was not changed until the maximum angle provided by the roller units.

\subsubsection{Rewards}
\label{subsubsec:rewards}
The RL agent accumulated a negative reward of -0.001 per step, while zero reward was added at subgoals and final goal (Figure \ref{fig:fig1}b). When the guidewire tip arrived at a terminal signal, a large negative reward of -0.5 was imposed. 

\begin{figure}
	\centering
    \includegraphics[scale = 0.5,keepaspectratio]{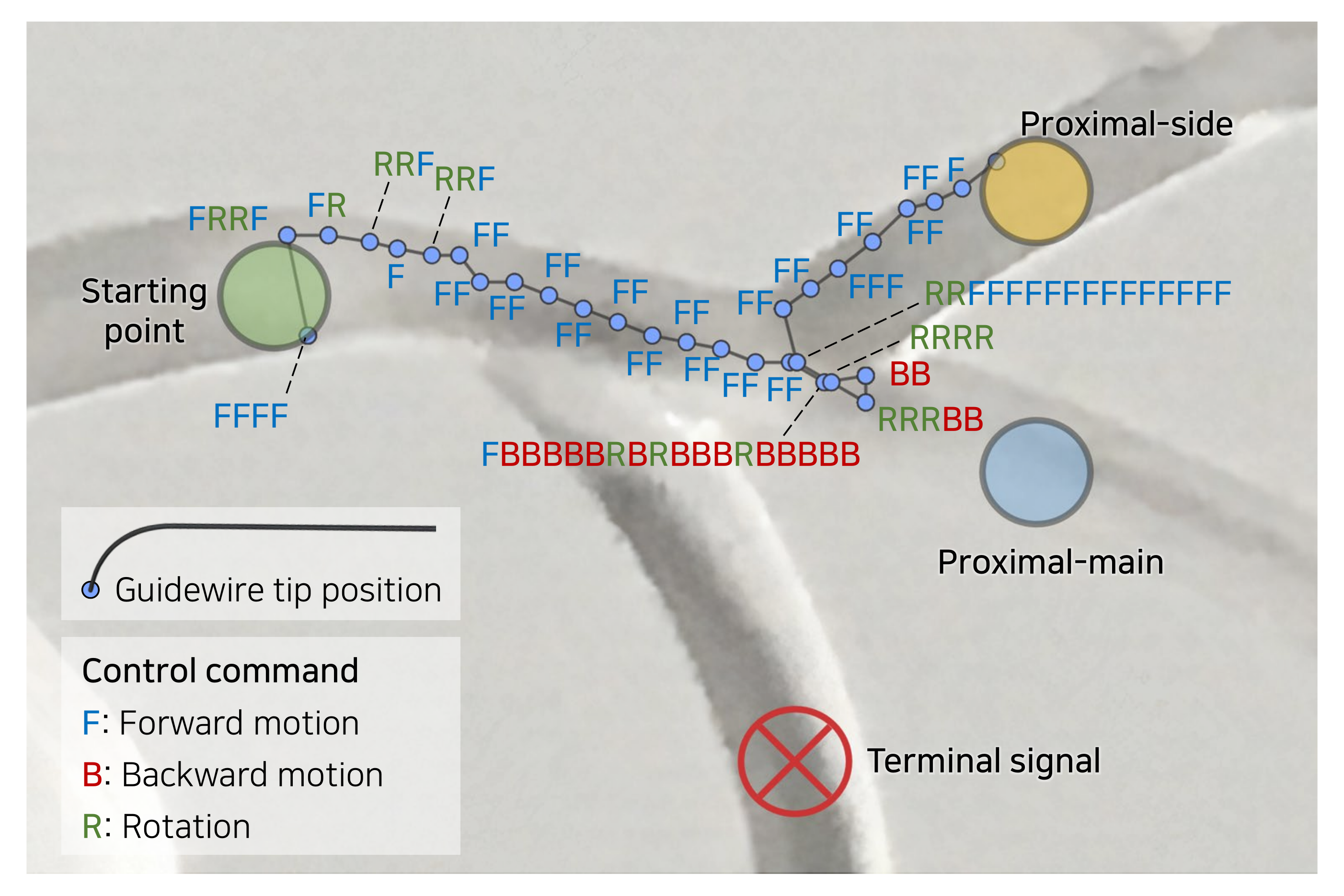}
	\caption{A trajectory of the guidewire tip in human operation as a demonstration. Even if given the same control command, the movement of the guidewire tip may vary depending on the tip orientation of the guidewire and the combination of control operations previously applied. For example, although four forward commands (FFFF) are consecutively selected at the starting point, the guidewire tip does not advance forward. Deformation of the guidewire caused by the friction forces on the walls of the catheter and the blood vessels causes non-linearity between the control command and the movement of the distal end, occasionally leaving the operator unable to predict the next state.}
	\label{fig:fig3}
\end{figure}

\subsubsection{Human demonstrations}
\label{subsubsec:human_demo}
DQfD (\citealp{hester2018deep}) was proposed to enhance the performance of reinforcement learning with a small amount of demonstration data. In our application of DQfD, human demonstrations were used to pre-train the network with supervised learning and were sampled with a high priority in the replay memory for reinforcement learning. To record demonstration data of experiments using the 2D phantom, 10 episodes per target location in the proximal and medial zones were created, as trained personnel generated discrete control commands using a keyboard (Figure \ref{fig:fig3}).

\section{Results}
\label{sec:results}

\subsection{Navigation performance in 2D phantom}
\label{subsec:2D}
First, the RL navigation in this study was aimed at delivering the guidewire to a target location at the main vessel or a side branch in a 2D phantom (PCI trainer for beginners, Medi Alpha Co., Ltd., Japan). The training was conducted by dividing the left anterior descending arteries into three parts and expanding the navigation area step by step (Figure \ref{fig:fig2}a). The RL performance was evaluated by independently performing the procedure three times per experiment, using a new guidewire each time. The target was selected at the beginning of every episode by uniform random distribution. A paired Wilcoxon test was used to compare the operation time and number of steps between RL agents. Values of p < 0.001 were considered statistically significant. Statistical analyses were performed using R package and SPSS 17.0 for Windows (IBM Corp., Armonk, NY, USA). 

\begin{table}
	\caption{Evaluation metrics for assessing the performance of RL agents.}
	\centering
	\begin{tabular}{cccccccccc}
		\toprule
		{ }&{ }&\multicolumn{2}{c}{Success rate}&\multicolumn{3}{c}{Number of steps}&\multicolumn{3}{c}{Operating time (s)}\\
		\cmidrule(r){3-10}

		{Exp.} &
		{Model} &
		{<95$\%$} &
		{<99$\%$} &
		{\begin{tabular}{@{}c@{}}First 100\\episodes\end{tabular}} &
		{\begin{tabular}{@{}c@{}}Last 100\\episodes\end{tabular}} &
		{\begin{tabular}{@{}c@{}}Last 500\\episodes\end{tabular}} &		
		{\begin{tabular}{@{}c@{}}First 100\\episodes\end{tabular}} &
		{\begin{tabular}{@{}c@{}}Last 100\\episodes\end{tabular}} &
		{\begin{tabular}{@{}c@{}}Last 500\\episodes\end{tabular}} \\
        
        \midrule
		\multirow{2}*{1} &
		{P1} &
		{201st} &
		{401st} &
		{\begin{tabular}{@{}c@{}}145.3\\$\pm$ 92.2\end{tabular}} &
		{\begin{tabular}{@{}c@{}}73.7\\$\pm$ 30.7\end{tabular}} &
		{\begin{tabular}{@{}c@{}}80.1\\$\pm$ 41.3\end{tabular}} &
		{\begin{tabular}{@{}c@{}}16.00\\$\pm$ 9.48\end{tabular}} &
		{\begin{tabular}{@{}c@{}}9.59\\$\pm$ 5.96\end{tabular}} &
		{\begin{tabular}{@{}c@{}}10.40\\$\pm$ 6.32\end{tabular}} \\

		\cmidrule(r){2-10}
        &
		{P2} &
		{188th} &
		{216th} &
		{\begin{tabular}{@{}c@{}}164.8\\$\pm$ 99.9\end{tabular}} &
		{\begin{tabular}{@{}c@{}}63.1\\$\pm$ 19.4\end{tabular}} &
		{\begin{tabular}{@{}c@{}}67.7\\$\pm$ 28.3\end{tabular}} &
		{\begin{tabular}{@{}c@{}}17.43\\$\pm$ 9.70\end{tabular}} &
		{\begin{tabular}{@{}c@{}}8.99\\$\pm$ 5.88\end{tabular}} &
		{\begin{tabular}{@{}c@{}}9.29\\$\pm$ 6.00\end{tabular}} \\

        \midrule
		\multirow{3}*{2} &
		{M1} &
		{560th} &
		{-} &
		{\begin{tabular}{@{}c@{}}179.3\\$\pm$ 85.5\end{tabular}} &
		{\begin{tabular}{@{}c@{}}203.2\\$\pm$ 77.7\end{tabular}} &
		{\begin{tabular}{@{}c@{}}203.8\\$\pm$ 77.0\end{tabular}} &
		{\begin{tabular}{@{}c@{}}18.94\\$\pm$ 9.19\end{tabular}} &
		{\begin{tabular}{@{}c@{}}20.43\\$\pm$ 9.08\end{tabular}} &
		{\begin{tabular}{@{}c@{}}20.78\\$\pm$ 9.26\end{tabular}} \\    
	
		\cmidrule(r){2-10}
        &
		{M2} &
		{195th} &
		{325th} &
		{\begin{tabular}{@{}c@{}}291.5\\$\pm$ 142.3\end{tabular}} &
		{\begin{tabular}{@{}c@{}}162.2\\$\pm$ 33.3\end{tabular}} &
		{\begin{tabular}{@{}c@{}}165.3\\$\pm$ 40.2\end{tabular}} &
		{\begin{tabular}{@{}c@{}}26.83\\$\pm$ 12.42\end{tabular}} &
		{\begin{tabular}{@{}c@{}}17.48\\$\pm$ 7.03\end{tabular}} &
		{\begin{tabular}{@{}c@{}}17.63\\$\pm$ 7.40\end{tabular}} \\       

		\cmidrule(r){2-10}
        &
		{M3} &
		{212th} &
		{446th} &
		{\begin{tabular}{@{}c@{}}435.9\\$\pm$ 116.3\end{tabular}} &
		{\begin{tabular}{@{}c@{}}157.6\\$\pm$ 37.2\end{tabular}} &
		{\begin{tabular}{@{}c@{}}160.7\\$\pm$ 41.3\end{tabular}} &
		{\begin{tabular}{@{}c@{}}39.41\\$\pm$ 11.64\end{tabular}} &
		{\begin{tabular}{@{}c@{}}17.24\\$\pm$ 6.66\end{tabular}} &
		{\begin{tabular}{@{}c@{}}17.56\\$\pm$ 7.17\end{tabular}} \\

        \midrule
		\multirow{2}*{3} &
		{D1} &
		{-} &
		{-} &
		{\begin{tabular}{@{}c@{}}346.6\\$\pm$ 154.5\end{tabular}} &
		{\begin{tabular}{@{}c@{}}375.6\\$\pm$ 180.8\end{tabular}} &
		{\begin{tabular}{@{}c@{}}353.9\\$\pm$ 186.4\end{tabular}} &
		{\begin{tabular}{@{}c@{}}30.16\\$\pm$ 12.57\end{tabular}} &
		{\begin{tabular}{@{}c@{}}33.63\\$\pm$ 16.00\end{tabular}} &
		{\begin{tabular}{@{}c@{}}32.46\\$\pm$ 16.72\end{tabular}} \\

		\cmidrule(r){2-10}
        &
		{D2} &
		{646th} &
		{-} &
		{\begin{tabular}{@{}c@{}}218.7\\$\pm$ 134.2\end{tabular}} &
		{\begin{tabular}{@{}c@{}}183.6\\$\pm$ 90.7\end{tabular}} &
		{\begin{tabular}{@{}c@{}}184.8\\$\pm$ 98.6\end{tabular}} &
		{\begin{tabular}{@{}c@{}}21.69\\$\pm$ 12.97\end{tabular}} &
		{\begin{tabular}{@{}c@{}}21.65\\$\pm$ 11.44\end{tabular}} &
		{\begin{tabular}{@{}c@{}}21.21\\$\pm$ 11.47\end{tabular}} \\		
		
		\bottomrule
	\end{tabular}
	\label{tab:table2}
\end{table}

\begin{figure}
	\centering
    \includegraphics[width=\textwidth,keepaspectratio]{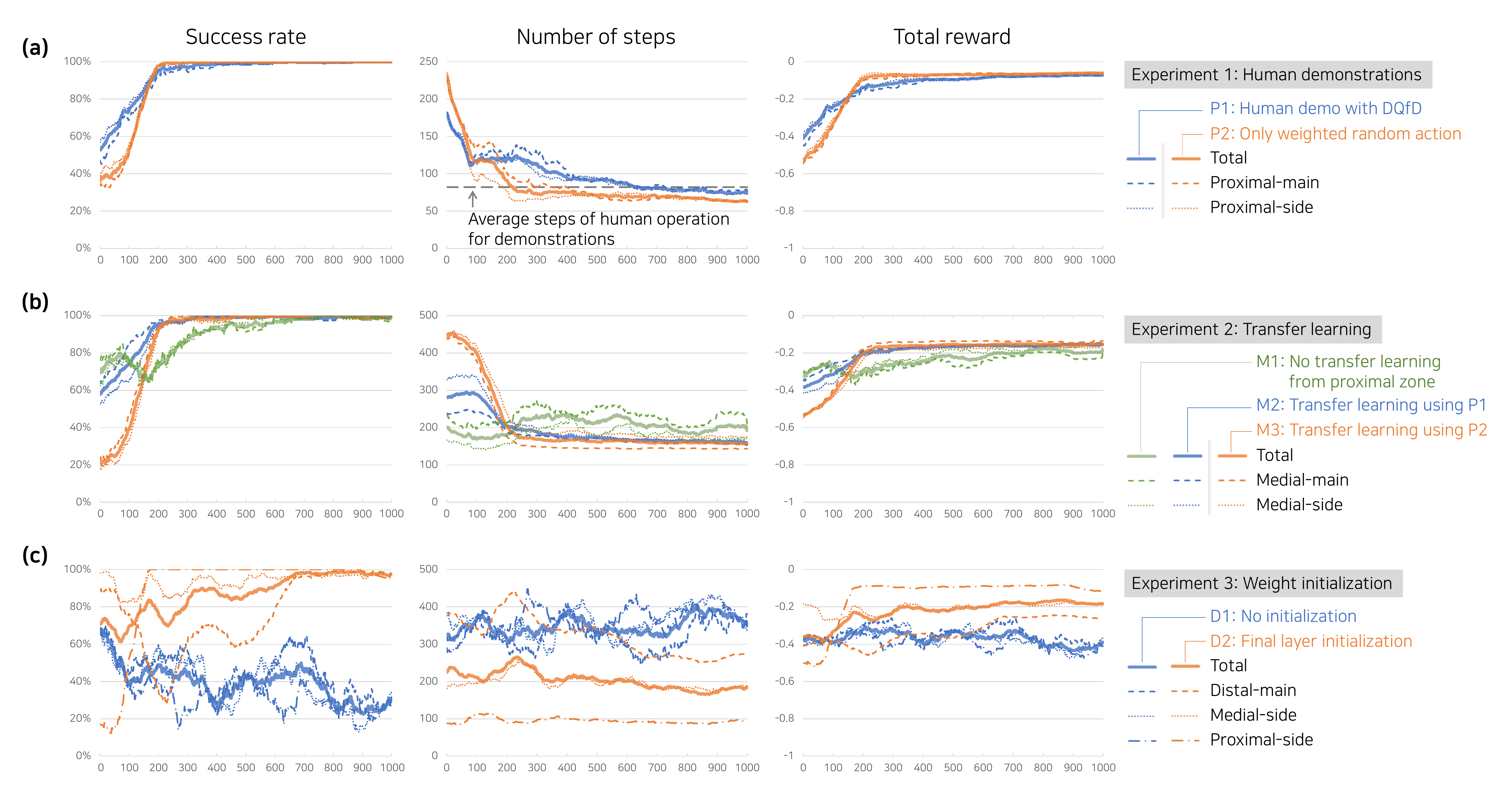}
	\caption{Navigation performance over episodes. (a) Human demonstrations with DQfD (\citealp{hester2018deep}) accelerate the training speed of RL agent, but require more episodes to succeed 100$\%$ of the time in the proximal zone. (b) Transfer learning strategy allows the RL agent to demonstrate the same level of training speed as the proximal zone. (c) When the final layer is initialized, the RL agent can make the guidewire reach the targets of different distances.}
	\label{fig:fig4}
\end{figure}

\begin{figure}
	\centering
    \includegraphics[width=\textwidth,keepaspectratio]{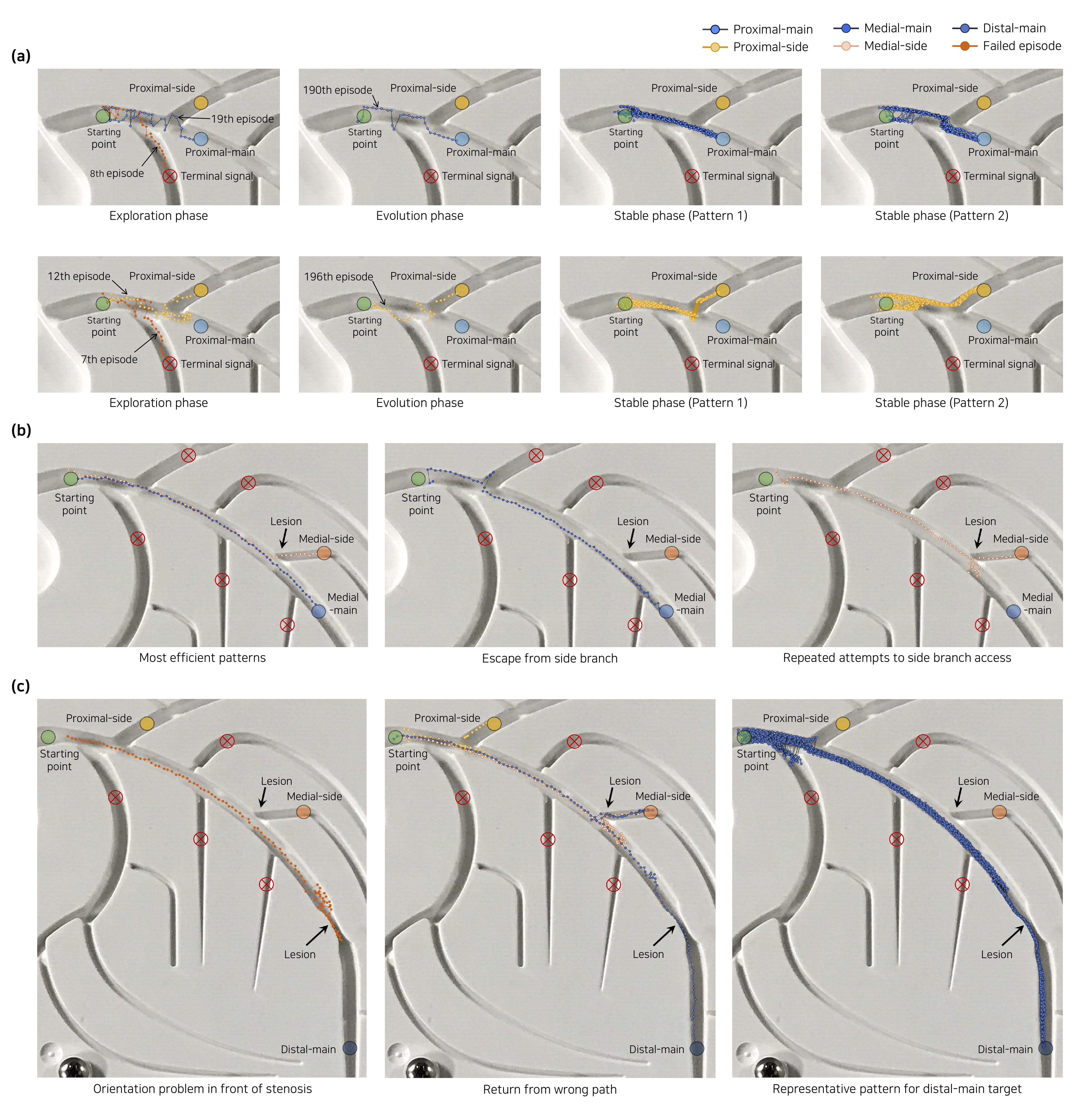}
	\caption{Tip trajectories of the guidewire in RL navigation. (a) In the proximal zone, the RL agent initially explores the path in a stochastic pattern, and as the RL evolves, the guidewire successfully reaches its goal by repeating efficient patterns in the stable phase. (b) In the medial zone, the RL agent finds an effective way to move the guidewire into a small side branch where the ostium is narrowed. (c) The RL agent passes through the severe obstruction in the distal zone by facing the guidewire tip to the right with respect to the travel direction.}
	\label{fig:fig5}
\end{figure}

\subsubsection{Proximal targets: Human demonstrations}
\label{subsubsec:prox}
For the main and side targets in the proximal zone, we evaluated the effects of human demonstrations with DQfD on RL performance for guidewire navigation (Figure \ref{fig:fig2}c). A stenotic lesion was located in the branching area between the targets, which hindered the guidewire control. Another side branch opposite to the proximal side target was set as a terminal signal.\\
When human demonstrations were added in the replay memory (P1 model), the learning speed was faster (<175 episodes in Figure \ref{fig:fig4}a) than the RL agent trained with only WRA (P2 model). In this case, the success rate of P2 model increased rapidly, reaching 99$\%$ first at 216th episode (Table \ref{tab:table2}). After 500 episodes, RL navigation hardly failed while P2 models required significantly less control commands (80.1 ± 41.3 vs. 67.7 ± 28.3, p < 0.0001) and reduced operating time (10.40 ± 6.32 s vs. 9.29 ± 6.00 s, p < 0.0001). Compared to the human operation for demonstration data (82.1 ± 34.2), the reduction rates in the number of steps were 10.3$\%$ and 23.1$\%$ for P1 and P2 models in the final 100 episodes, respectively. For both models, although the success rate for the proximal-main target (vs. proximal-side target) was slightly lower in the beginning, as the RL training progresses, the difference between the navigation goals almost disappeared.\\
In the early stage of training, the RL agent explored the path in a stochastic pattern (exploration phase in Figure \ref{fig:fig5}a). As the training progressed, unnecessary changes in the orientation of the guidewire tip were gradually reduced, and the probability of escape from untargeted branches was improved (evolving phase). In ‘stable phase’ at the last stage of training, two representative patterns were found in the trajectories of guidewire navigation. The first pattern was that after proceeding along the centerline of the main vessel, the guidewire rotated sharply to the proximal-side target in the bifurcation area or advanced to the proximal-main target. The second pattern was characterized by avoiding the branch vessel of terminal signal. Then, the RL agent steered the guidewire along the sidewall of the side branch (proximal-side target) or used the evasive movement again to the opposite side (proximal-main target).

\subsubsection{Medial targets: Transfer learning}
\label{subsubsec:med}
Because the travel distances to the medial targets were roughly three times longer than those to the proximal targets, it was extremely difficult to reach the goals with only WRA, especially the medial-side target (Figure \ref{fig:fig2}a). For the medial targets, the transfer learning approach was applied by initializing the network using the trained models in the proximal zone. The M1 model, as a control, was trained using human demonstration with DQfD like the P1 model. The M2 model brought the initial weights from the P1 model, which generated the initial transitions to the medial targets. The initial transitions of the M3 model using the weight of the P2 model were produced with both the transferred weight and WRA (Figure \ref{fig:fig2}c and Table \ref{tab:table1}).\\
Despite the increased travel distance with intervening multiple branches, the success rate of M2 and M3 models increased sharply in the same pattern as the proximal experiment, indicating > 95$\%$ from the 212th episode for both the models (Figure \ref{fig:fig4}b and Table \ref{tab:table2}). After the success rate became saturated, little variation was found between the performance of the two models using transfer learning. Also, the number of steps and total reward per episode remained stable in terms of averaging of 100 episodes. Although human demonstrations allowed the RL agent to temporarily produce better results (M1 model), after the initial stage (> 200 episodes), the performance of M2 and M3 models exceeded the RL agent trained without transfer learning.\\
The most efficient patterns for medial targets, obviously, were to follow the centerline primarily using forward commands, which accounted for most of the last half of the experiment (Figure \ref{fig:fig5}b). The main reasons for the failure or longer travel distance in the navigation were that the guidewire was misled to the side branches (medial-main target) and the orientation of the guidewire tip had to be changed repeatedly to pass through the narrowing (medial-side target).

\subsubsection{Distal-main and side targets: Weight initialization}
\label{subsubsec:dist}
The goal was to demonstrate that the RL agent was viable for the major destinations of guidewire navigation: proximal-side, medial-side, and distal-main targets. Transfer learning was applied using the trained weights of the M3 model. To address the overfitting issue, the weight initialization was applied by replacing the final layer of the convolutional neural network with randomly generated parameters.\\
Without weight initialization (D1 model), the performance of the RL agent regressed as the training continued. For the last 100 episodes, D1 model produced a success rate of 25$\%$ and mostly failed to reach the distal-main target (Figure \ref{fig:fig4}c). When the final layer initialization was applied (D2 model), the success rate for the medial-side and distal-main targets fluctuated, but eventually approached 100$\%$ for all targets. The success rate of the D2 model was 98.0$\%$ in average for the last 300 episodes. The change in the number of steps in the training process was relatively small compared to the proximal and medial zones, because the navigation was terminated early in the failed episodes.\\
Initially, the guidewire control suffered from the orientation adjustment of the guidewire in front of the stenosis (Figure \ref{fig:fig5}c). Unless the guidewire tip faced to the right relative to the travel direction, it was exceedingly difficult to proceed through the distal obstruction. As the training progressed, even when the guidewire moved along an incorrect route, the RL agent reverted it to the path that could reach the designated goal as it learned in the previous experimental stages. Eventually, the navigation trajectories for the distal-main target almost converged except for irregular patterns around the largest side branches in the proximal zone.\\

\begin{figure}
	\centering
    \includegraphics[width=\textwidth,keepaspectratio]{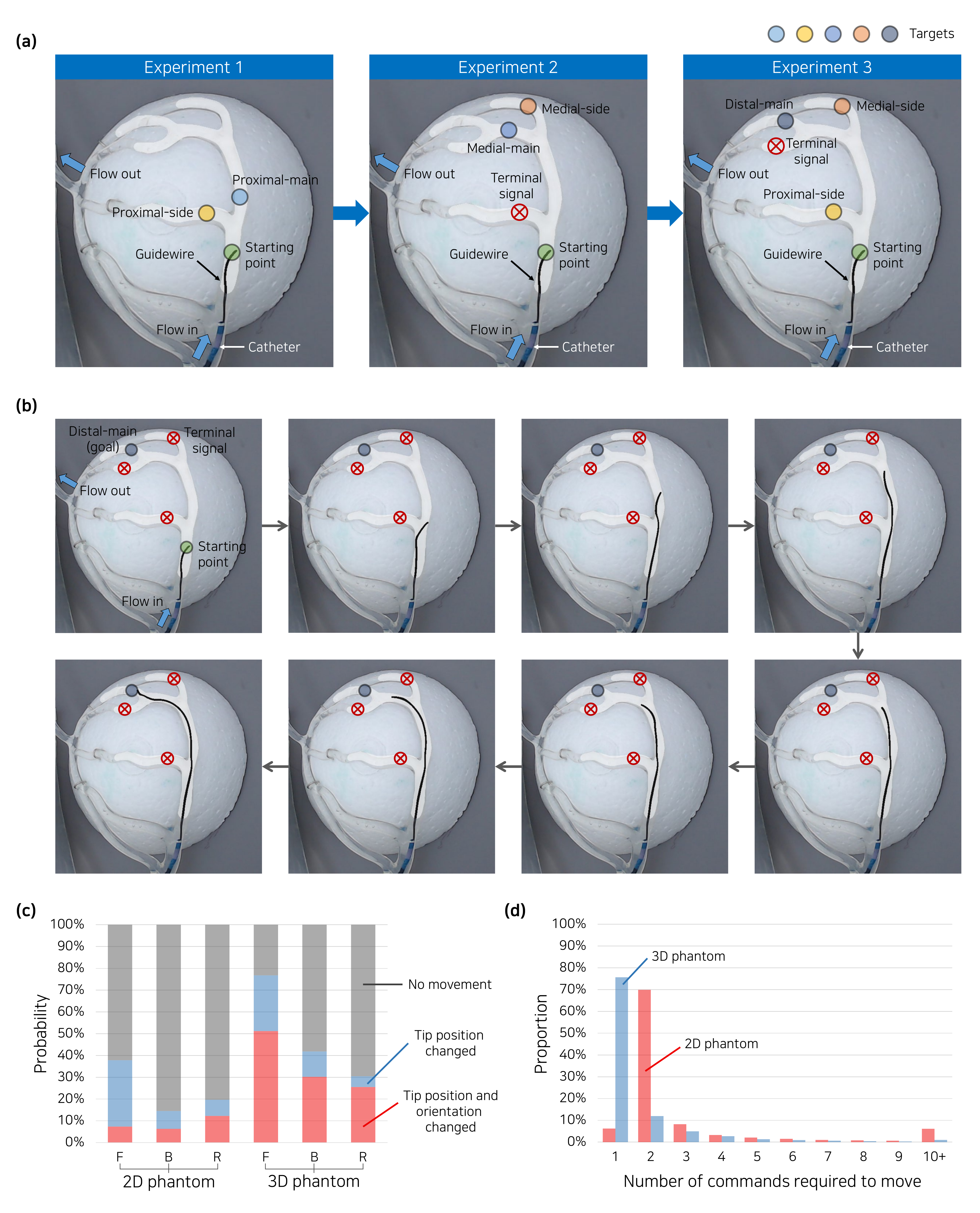}
	\caption{(a) Navigation targets in 3D phantom. The target names may differ from the clinical standard. (b) Snapshots of RL navigation to the distal-main target. (c) Probability of guidewire movement corresponding to forward (F), backward (B) and rotation (R) commands. (d) Number of control commands required for guidewire movement.}
	\label{fig:fig6}
\end{figure}

\subsection{Validation in 3D phantom}
\label{subsec:3D}
The training process of our framework was further validated using a 3D phantom. Expanding the navigation area in the 3D phantom (Embedded Coronary Model, Trandomed 3D Medical Technology Co., Ltd., China), the training methods of the P2, M3, and D2 models were applied sequentially following the best scenario in the 2D phantom. In the 3D experimental setup (Figure \ref{fig:fig6}a), the vessel and the guidewire placed away from the center of the camera view could be detected as shorter than the actual length, like the shortening effect in X-ray coronary angiography. Also, fluid at physiological flow rate was supplied from an output-adjustable pump (WT300-1JA, Longer Precision Pump Co., Ltd., China) to the right coronary artery (RCA), which could affect the dynamic behavior of the guidewire along with the silicon wall.\\
Despite substantial changes in the experimental environment and thereby mutual interaction (Figures \ref{fig:fig6}c and \ref{fig:fig6}d), an RL agent was constructed that was able to steer the guidewire to reach all valid targets (see Figure \ref{fig:figA} in Appendix). The navigation performance of the RL agent improved through the exploration and evolution phases, and eventually, it became capable of maneuvering the guidewire into the vessels with wavy walls and differing branching patterns (Figure \ref{fig:fig6}b and Video A1).

\section{Discussion}
\label{sec:dicussion}
Our framework demonstrated the potential applicability of autonomous guidewire navigation using RL. To accelerate the training speed and avoid mechanical fatigue of the guidewire, human demonstrations with DQfD (\citealp{hester2018deep}), transfer learning, and weight initialization were evaluated as a segment-wise learning approach. The focus window and subgoals were introduced to customize the state and reward for RL agent, respectively. The RL agent improved navigation performance through ‘exploration’ and ‘evolution’ phases, which eventually enabled the guidewire to reach all valid targets in a ‘stable’ phase.\\
Human demonstrations initially accelerated the training speed but required more time to further increase the success rate. Sampling human demonstrations with a high priority, even after the RL agent's performance exceeds the human demonstrations, may be a hindrance to performance improvement. The patterns of the human demonstrations were suboptimal, which rarely appeared in the ‘stable’ phase (\citealp{gao2018reinforcement}). Also, the difference in the input frequency between the human operator and the RL agent may cause different interactions between the guidewire tip and the experimental environment. In the navigation to the medial and distal targets, the segment-wise approach helped to collect better transitions for the training of the RL agent for a target with a low probability of reaching it only by random action. The transfer learning, which is commonly used for learning speed and performance of deep networks (\citealp{girshick2014rich}), also contributed to reduce the time required for 100$\%$ success.\\
In realistic physiological conditions, providing accurate state information to the RL agent is the key to applying RL in three-dimensional deformation of the vascular pathways with a living heartbeat. Uncertainties in registration can be an obstacle to apply our framework that required precise position information to define ‘state’ and apply subgoals. To detect the relative location of the guidewire within the coronary tree, a dynamic coronary roadmap can be helpful (\citealp{piayda2018dynamic}). The latest updated method provides a real-time registration of X-ray angiography with the guidewire tip in fluoroscopic images using ECG gating (\citealp{ma2020dynamic}). Also, deep-learning segmentation of major vessels in X-ray angiography offers rapid and accurate identification of the target vessel to be reached (\citealp{yang2019deep}).\\
The ultimate goal of autonomous navigation using deep RL is to build a generalized model encompassing the anatomical diversity of the coronary arteries. Time and cost for training can be an obstacle that fundamentally limits the application of RL navigation. To this end, the development of novel simulators is essential (\citealp{wang2017robust}; \citealp{dulac2019challenges}). Virtual simulators provide an opportunity to train more subjects quickly at a low cost. Distributed RL can improve the training and the performance of the RL agent using the strength of virtual simulators (\citealp{mnih2016asynchronous}). Advancements in the modeling of cardiovascular anatomy (\citealp{corral2020digital}) and interventional devices (\citealp{sharei2018navigation}) support the construction of virtual simulators that more accurately mimic interactions in the human body. Physical simulators may help translate from virtual simulators to in-vivo applications by relieving safety issues. Integration of novel 3D printing techniques (\citealp{stepniak2020novel}) with functional modeling of the cardiovascular system (\citealp{vukicevic2017cardiac}) may allow implementation of dynamic response in interventional devices in physical simulators. Our framework is expected to contribute to the adoption of autonomous navigation not only by providing data necessary for modeling virtual simulators, but also by presenting guidance on training methods for physical simulators involving mechanical fatigue.

\bibliographystyle{unsrtnat}
\bibliography{references.bib}

\begin{thebibliography}{43}
\providecommand{\natexlab}[1]{#1}
\providecommand{\url}[1]{\texttt{#1}}
\expandafter\ifx\csname urlstyle\endcsname\relax
  \providecommand{\doi}[1]{doi: #1}\else
  \providecommand{\doi}{doi: \begingroup \urlstyle{rm}\Url}\fi

\bibitem[Dodge~Jr et~al.(1992)Dodge~Jr, Brown, Bolson, and
  Dodge]{dodge1992lumen}
J~Theodore Dodge~Jr, B~Greg Brown, Edward~L Bolson, and Harold~T Dodge.
\newblock Lumen diameter of normal human coronary arteries.
  \uppercase{I}nfluence of age, sex, anatomic variation, and left ventricular
  hypertrophy or dilation.
\newblock \emph{Circulation}, 86\penalty0 (1):\penalty0 232--246, 1992.

\bibitem[Guttmann et~al.(2017)Guttmann, Jones, Gulati, Kotecha, Fayed, Patel,
  Crake, Ozkor, Wragg, Smith, Weerackody, Knight, Mathur, and
  O’Mahony]{guttmann2017prevalence}
Oliver~P Guttmann, Daniel~A Jones, Ankur Gulati, Tushar Kotecha, Hossam Fayed,
  Deven Patel, Tom Crake, Mick Ozkor, Andrew Wragg, Elliot~J Smith, Roshan
  Weerackody, Charles~J Knight, Anthony Mathur, and Constantinos O’Mahony.
\newblock Prevalence and outcomes of coronary artery perforation during
  percutaneous coronary intervention.
\newblock \emph{EuroIntervention}, 13\penalty0 (5):\penalty0 e595--e601, 2017.

\bibitem[Beyar et~al.(2005)Beyar, Wenderow, Lindner, Kumar, and
  Shofti]{beyar2005concept}
Rafael Beyar, Tal Wenderow, Doron Lindner, Ganesh Kumar, and Rona Shofti.
\newblock Concept, design and pre-clinical studies for remote control
  percutaneous coronary interventions.
\newblock \emph{EuroIntervention}, 1\penalty0 (3):\penalty0 340--345, 2005.

\bibitem[Kiemeneij et~al.(2008)Kiemeneij, Patterson, Amoroso, Laarman, and
  Slagboom]{kiemeneij2008use}
Ferdinand Kiemeneij, Mark~S Patterson, Giovanni Amoroso, GertJan Laarman, and
  Ton Slagboom.
\newblock Use of the stereotaxis niobe{\textregistered} magnetic navigation
  system for percutaneous coronary intervention: Results from 350 consecutive
  patients.
\newblock \emph{Catheterization and Cardiovascular Interventions}, 71\penalty0
  (4):\penalty0 510--516, 2008.

\bibitem[Weisz et~al.(2013)Weisz, Metzger, Caputo, Delgado, Marshall, Vetrovec,
  Reisman, Waksman, Granada, Novack, Moses, and Carrozza]{weisz2013safety}
Giora Weisz, D~Christopher Metzger, Ronald~P Caputo, Juan~A Delgado, J~Jeffrey
  Marshall, George~W Vetrovec, Mark Reisman, Ron Waksman, Juan~F Granada,
  Victor Novack, Jeffrey~W. Moses, and Joseph~P. Carrozza.
\newblock Safety and feasibility of robotic percutaneous coronary intervention:
  \uppercase{PRECISE} (\uppercase{P}ercutaneous
  \uppercase{R}obotically-\uppercase{E}nhanced \uppercase{C}oronary
  \uppercase{I}ntervention) study.
\newblock \emph{Journal of the American College of Cardiology}, 61\penalty0
  (15):\penalty0 1596--1600, 2013.

\bibitem[Patel et~al.(2020)Patel, Shah, Soni, Radadiya, Patel, Tiwari, and
  Pancholy]{patel2020comparison}
Tejas~M Patel, Sanjay~C Shah, Yash~Y Soni, Rajni~C Radadiya, Gaurav~A Patel,
  Pradyot~O Tiwari, and Samir~B Pancholy.
\newblock Comparison of robotic percutaneous coronary intervention with
  traditional percutaneous coronary intervention: A propensity score--matched
  analysis of a large cohort.
\newblock \emph{Circulation: Cardiovascular Interventions}, 13\penalty0
  (5):\penalty0 e008888, 2020.

\bibitem[Hirai et~al.(2020)Hirai, Kearney, Kataruka, Gosch, Brandt, Nicholson,
  Lombardi, Grantham, and Salisbury]{hirai2020initial}
Taishi Hirai, Kathleen Kearney, Akash Kataruka, Kensey~L Gosch, Hunter Brandt,
  William~J Nicholson, William~L Lombardi, J~Aaron Grantham, and Adam~C
  Salisbury.
\newblock Initial report of safety and procedure duration of robotic-assisted
  chronic total occlusion coronary intervention.
\newblock \emph{Catheterization and Cardiovascular Interventions}, 95\penalty0
  (1):\penalty0 165--169, 2020.

\bibitem[Mahmud et~al.(2017)Mahmud, Naghi, Ang, Harrison, Behnamfar,
  Pourdjabbar, Reeves, and Patel]{mahmud2017demonstration}
Ehtisham Mahmud, Jesse Naghi, Lawrence Ang, Jonathan Harrison, Omid Behnamfar,
  Ali Pourdjabbar, Ryan Reeves, and Mitul Patel.
\newblock Demonstration of the safety and feasibility of robotically assisted
  percutaneous coronary intervention in complex coronary lesions: results of
  the \uppercase{CORA-PCI} study (\uppercase{C}omplex \uppercase{R}obotically
  \uppercase{A}ssisted \uppercase{P}ercutaneous \uppercase{C}oronary
  \uppercase{I}ntervention).
\newblock \emph{JACC: Cardiovascular Interventions}, 10\penalty0 (13):\penalty0
  1320--1327, 2017.

\bibitem[Patel et~al.(2019)Patel, Shah, and Pancholy]{patel2019long}
Tejas~M Patel, Sanjay~C Shah, and Samir~B Pancholy.
\newblock Long distance tele-robotic-assisted percutaneous coronary
  intervention: A report of first-in-human experience.
\newblock \emph{EClinicalMedicine}, 14:\penalty0 53--58, 2019.

\bibitem[Attanasio et~al.(2021)Attanasio, Scaglioni, De~Momi, Fiorini, and
  Valdastri]{attanasio2021autonomy}
Aleks Attanasio, Bruno Scaglioni, Elena De~Momi, Paolo Fiorini, and Pietro
  Valdastri.
\newblock Autonomy in surgical robotics.
\newblock \emph{Annual Review of Control, Robotics, and Autonomous Systems},
  4:\penalty0 651--679, 2021.

\bibitem[Zemmar et~al.(2020)Zemmar, Lozano, and Nelson]{zemmar2020rise}
Ajmal Zemmar, Andres~M Lozano, and Bradley~J Nelson.
\newblock The rise of robots in surgical environments during
  \uppercase{COVID}-19.
\newblock \emph{Nature Machine Intelligence}, 2\penalty0 (10):\penalty0
  566--572, 2020.

\bibitem[Sardar et~al.(2019)Sardar, Abbott, Kundu, Aronow, Granada, and
  Giri]{sardar2019impact}
Partha Sardar, J~Dawn Abbott, Amartya Kundu, Herbert~D Aronow, Juan~F Granada,
  and Jay Giri.
\newblock Impact of artificial intelligence on interventional cardiology: From
  decision-making aid to advanced interventional procedure assistance.
\newblock \emph{JACC: Cardiovascular Interventions}, 12\penalty0 (14):\penalty0
  1293--1303, 2019.

\bibitem[Silver et~al.(2016)Silver, Huang, Maddison, Guez, Sifre, van~den
  Driessche, Schrittwieser, Antonoglou, Panneershelvam, Lanctot, Dieleman,
  Grewe, Nham, Kalchbrenner, Sutskever, Lillicrap, Leach, Kavukcuoglu, Graepel,
  and Hassabis]{silver2016mastering}
David Silver, Aja Huang, Chris~J Maddison, Arthur Guez, Laurent Sifre, George
  van~den Driessche, Julian Schrittwieser, Ioannis Antonoglou, Veda
  Panneershelvam, Marc Lanctot, Sander Dieleman, Dominik Grewe, John Nham, Nal
  Kalchbrenner, Ilya Sutskever, Timothy Lillicrap, Madeleine Leach, Koray
  Kavukcuoglu, Thore Graepel, and Demis Hassabis.
\newblock Mastering the game of \uppercase{G}o with deep neural networks and
  tree search.
\newblock \emph{Nature}, 529\penalty0 (7587):\penalty0 484--489, 2016.

\bibitem[Mnih et~al.(2015)Mnih, Kavukcuoglu, Silver, Rusu, Veness, Bellemare,
  Graves, Riedmiller, Fidjeland, Ostrovski, Petersen, Beattie, Sadik,
  Antonoglou, King, Kumaran, Wierstra, Legg, and Hassabis]{mnih2015human}
Volodymyr Mnih, Koray Kavukcuoglu, David Silver, Andrei~A Rusu, Joel Veness,
  Marc~G Bellemare, Alex Graves, Martin Riedmiller, Andreas~K Fidjeland, Georg
  Ostrovski, Stig Petersen, Charles Beattie, Amir Sadik, Ioannis Antonoglou,
  Helen King, Dharshan Kumaran, Daan Wierstra, Shane Legg, and Demis Hassabis.
\newblock Human-level control through deep reinforcement learning.
\newblock \emph{Nature}, 518\penalty0 (7540):\penalty0 529--533, 2015.

\bibitem[Levine et~al.(2016)Levine, Finn, Darrell, and Abbeel]{levine2016end}
Sergey Levine, Chelsea Finn, Trevor Darrell, and Pieter Abbeel.
\newblock End-to-end training of deep visuomotor policies.
\newblock \emph{The Journal of Machine Learning Research}, 17\penalty0
  (1):\penalty0 1334--1373, 2016.

\bibitem[Gandhi et~al.(2017)Gandhi, Pinto, and Gupta]{gandhi2017learning}
Dhiraj Gandhi, Lerrel Pinto, and Abhinav Gupta.
\newblock Learning to fly by crashing.
\newblock In \emph{2017 IEEE/RSJ International Conference on Intelligent Robots
  and Systems (IROS)}, pages 3948--3955. IEEE, 2017.

\bibitem[Pinto et~al.(2017)Pinto, Andrychowicz, Welinder, Zaremba, and
  Abbeel]{pinto2017asymmetric}
Lerrel Pinto, Marcin Andrychowicz, Peter Welinder, Wojciech Zaremba, and Pieter
  Abbeel.
\newblock Asymmetric actor critic for image-based robot learning.
\newblock \emph{arXiv preprint arXiv:1710.06542}, 2017.

\bibitem[You et~al.(2019)You, Bae, Moon, Kweon, and Choi]{you2019automatic}
Hyeonseok You, EunKyung Bae, Youngjin Moon, Jihoon Kweon, and Jaesoon Choi.
\newblock Automatic control of cardiac ablation catheter with deep
  reinforcement learning method.
\newblock \emph{Journal of Mechanical Science and Technology}, 33\penalty0
  (11):\penalty0 5415--5423, 2019.

\bibitem[Behr et~al.(2019)Behr, Pusch, Siegfarth, H{\"u}sener, M{\"o}rschel,
  and Karstensen]{behr2019deep}
Tobias Behr, Tim~Philipp Pusch, Marius Siegfarth, Dominik H{\"u}sener, Tobias
  M{\"o}rschel, and Lennart Karstensen.
\newblock Deep reinforcement learning for the navigation of neurovascular
  catheters.
\newblock \emph{Current Directions in Biomedical Engineering}, 5\penalty0
  (1):\penalty0 5--8, 2019.

\bibitem[Karstensen et~al.(2020)Karstensen, Behr, Pusch, Mathis-Ullrich, and
  Stallkamp]{karstensen2020autonomous}
Lennart Karstensen, Tobias Behr, Tim~Philipp Pusch, Franziska Mathis-Ullrich,
  and Jan Stallkamp.
\newblock Autonomous guidewire navigation in a two dimensional vascular
  phantom.
\newblock \emph{Current Directions in Biomedical Engineering}, 6\penalty0 (1),
  2020.

\bibitem[Chi et~al.(2020)Chi, Dagnino, Kwok, Nguyen, Kundrat, Abdelaziz, Riga,
  Bicknell, and Yang]{chi2020collaborative}
Wenqiang Chi, Giulio Dagnino, Trevor~MY Kwok, Anh Nguyen, Dennis Kundrat,
  Mohamed~EMK Abdelaziz, Celia Riga, Colin Bicknell, and Guang-Zhong Yang.
\newblock Collaborative robot-assisted endovascular catheterization with
  generative adversarial imitation learning.
\newblock In \emph{2020 IEEE International Conference on Robotics and
  Automation (ICRA)}, pages 2414--2420. IEEE, 2020.

\bibitem[Zhao et~al.(2019)Zhao, Guo, Wang, Cui, Ma, Zeng, Liu, Jiang, Li, Shi,
  and Xiao]{zhao2019cnn}
Yan Zhao, Shuxiang Guo, Yuxin Wang, Jinxin Cui, Youchun Ma, Yuwen Zeng, Xinke
  Liu, Yuhua Jiang, Youxinag Li, Liwei Shi, and Nan Xiao.
\newblock A \uppercase{CNN}-based prototype method of unstructured surgical
  state perception and navigation for an endovascular surgery robot.
\newblock \emph{Medical \& Biological Engineering \& Computing}, 57\penalty0
  (9):\penalty0 1875--1887, 2019.

\bibitem[Hessel et~al.(2018)Hessel, Modayil, van Hasselt, Schaul, Ostrovski,
  Dabney, Horgan, Piot, Azar, and Silver]{hessel2018rainbow}
Matteo Hessel, Joseph Modayil, Hado van Hasselt, Tom Schaul, Georg Ostrovski,
  Will Dabney, Dan Horgan, Bilal Piot, Mohammad Azar, and David Silver.
\newblock Rainbow: Combining improvements in deep reinforcement learning.
\newblock In \emph{Thirty-second AAAI conference on artificial intelligence},
  2018.

\bibitem[van Hasselt et~al.(2016)van Hasselt, Guez, and Silver]{van2016deep}
Hado van Hasselt, Arthur Guez, and David Silver.
\newblock Deep reinforcement learning with double q-learning.
\newblock In \emph{Proceedings of the AAAI conference on artificial
  intelligence}, volume~30, 2016.

\bibitem[Schaul et~al.(2015)Schaul, Quan, Antonoglou, and
  Silver]{schaul2015prioritized}
Tom Schaul, John Quan, Ioannis Antonoglou, and David Silver.
\newblock Prioritized experience replay.
\newblock \emph{arXiv preprint arXiv:1511.05952}, 2015.

\bibitem[Wang et~al.(2016)Wang, Schaul, Hessel, Hasselt, Lanctot, and
  Freitas]{wang2016dueling}
Ziyu Wang, Tom Schaul, Matteo Hessel, Hado Hasselt, Marc Lanctot, and Nando
  Freitas.
\newblock Dueling network architectures for deep reinforcement learning.
\newblock In \emph{International conference on machine learning}, pages
  1995--2003. PMLR, 2016.

\bibitem[Fortunato et~al.(2017)Fortunato, Azar, Piot, Menick, Osband, Graves,
  Mnih, Munos, Hassabis, Pietquin, Blundell, and Legg]{fortunato2017noisy}
Meire Fortunato, Mohammad~Gheshlaghi Azar, Bilal Piot, Jacob Menick, Ian
  Osband, Alex Graves, Vlad Mnih, Remi Munos, Demis Hassabis, Olivier Pietquin,
  Charles Blundell, and Shane Legg.
\newblock Noisy networks for exploration.
\newblock \emph{arXiv preprint arXiv:1706.10295}, 2017.

\bibitem[Bellemare et~al.(2017)Bellemare, Dabney, and
  Munos]{bellemare2017distributional}
Marc~G Bellemare, Will Dabney, and R{\'e}mi Munos.
\newblock A distributional perspective on reinforcement learning.
\newblock In \emph{International Conference on Machine Learning}, pages
  449--458. PMLR, 2017.

\bibitem[Church et~al.(2020)Church, Lloyd, Hadsell, and Lepora]{church2020deep}
Alex Church, John Lloyd, Raia Hadsell, and Nathan~F Lepora.
\newblock Deep reinforcement learning for tactile robotics: Learning to type on
  a braille keyboard.
\newblock \emph{IEEE Robotics and Automation Letters}, 5\penalty0 (4):\penalty0
  6145--6152, 2020.

\bibitem[Hester et~al.(2018)Hester, Vecerik, Pietquin, Lanctot, Schaul, Piot,
  Horgan, Quan, Sendonaris, Osband, Agapiou, Leibo, and
  Gruslys]{hester2018deep}
Todd Hester, Matej Vecerik, Olivier Pietquin, Marc Lanctot, Tom Schaul, Bilal
  Piot, Dan Horgan, John Quan, Andrew Sendonaris, Ian Osband, John Agapiou,
  Joel~Z Leibo, and Audrunas Gruslys.
\newblock Deep q-learning from demonstrations.
\newblock In \emph{Thirty-second AAAI conference on artificial intelligence},
  2018.

\bibitem[Graves et~al.(2017)Graves, Bellemare, Menick, Munos, and
  Kavukcuoglu]{graves2017automated}
Alex Graves, Marc~G Bellemare, Jacob Menick, Remi Munos, and Koray Kavukcuoglu.
\newblock Automated curriculum learning for neural networks.
\newblock In \emph{international conference on machine learning}, pages
  1311--1320. PMLR, 2017.

\bibitem[Gao et~al.(2018)Gao, Xu, Lin, Yu, Levine, and
  Darrell]{gao2018reinforcement}
Yang Gao, Huazhe Xu, Ji~Lin, Fisher Yu, Sergey Levine, and Trevor Darrell.
\newblock Reinforcement learning from imperfect demonstrations.
\newblock \emph{arXiv preprint arXiv:1802.05313}, 2018.

\bibitem[Girshick et~al.(2014)Girshick, Donahue, Darrell, and
  Malik]{girshick2014rich}
Ross Girshick, Jeff Donahue, Trevor Darrell, and Jitendra Malik.
\newblock Rich feature hierarchies for accurate object detection and semantic
  segmentation.
\newblock In \emph{2014 IEEE conference on computer vision and pattern
  recognition (CVPR)}, pages 580--587, 2014.

\bibitem[Piayda et~al.(2018)Piayda, Kleinebrecht, Afzal, Bullens, Ter~Horst,
  Polzin, Veulemans, Dannenberg, Wimmer, Jung, Bönner, Kelm, Hellhammer, and
  Zeus]{piayda2018dynamic}
Kerstin Piayda, Laura Kleinebrecht, Shazia Afzal, Roland Bullens, Iris
  Ter~Horst, Amin Polzin, Verena Veulemans, Lisa Dannenberg, Anna~Christina
  Wimmer, Christian Jung, Florian Bönner, Malte Kelm, Katharina Hellhammer,
  and Tobias Zeus.
\newblock Dynamic coronary roadmapping during percutaneous coronary
  intervention: A feasibility study.
\newblock \emph{European journal of medical research}, 23\penalty0
  (1):\penalty0 1--7, 2018.

\bibitem[Ma et~al.(2020)Ma, Smal, Daemen, and van Walsum]{ma2020dynamic}
Hua Ma, Ihor Smal, Joost Daemen, and Theo van Walsum.
\newblock Dynamic coronary roadmapping via catheter tip tracking in
  \uppercase{X}-ray fluoroscopy with deep learning based bayesian filtering.
\newblock \emph{Medical image analysis}, 61:\penalty0 101634, 2020.

\bibitem[Yang et~al.(2019)Yang, Kweon, Roh, Lee, Kang, Park, Kim, Yang, Hur,
  Kang, Lee, Ahn, Kang, Park, Lee, Kim, Lee, Park, and Park]{yang2019deep}
Su~Yang, Jihoon Kweon, Jae-Hyung Roh, Jae-Hwan Lee, Heejun Kang, Lae-Jeong
  Park, Dong~Jun Kim, Hyeonkyeong Yang, Jaehee Hur, Do-Yoon Kang, Pil~Hyung
  Lee, Jung-Min Ahn, Soo-Jin Kang, Duk-Woo Park, Seung-Whan Lee, Young-Hak Kim,
  Cheol~Whan Lee, Seong-Wook Park, and Seung-Jung Park.
\newblock Deep learning segmentation of major vessels in \uppercase{X}-ray
  coronary angiography.
\newblock \emph{Scientific reports}, 9\penalty0 (1):\penalty0 1--11, 2019.

\bibitem[Wang et~al.(2017)Wang, Merel, Reed, Wayne, de~Freitas, and
  Heess]{wang2017robust}
Ziyu Wang, Josh Merel, Scott Reed, Greg Wayne, Nando de~Freitas, and Nicolas
  Heess.
\newblock Robust imitation of diverse behaviors.
\newblock \emph{arXiv preprint arXiv:1707.02747}, 2017.

\bibitem[Dulac-Arnold et~al.(2019)Dulac-Arnold, Mankowitz, and
  Hester]{dulac2019challenges}
Gabriel Dulac-Arnold, Daniel Mankowitz, and Todd Hester.
\newblock Challenges of real-world reinforcement learning.
\newblock \emph{arXiv preprint arXiv:1904.12901}, 2019.

\bibitem[Mnih et~al.(2016)Mnih, Badia, Mirza, Graves, Lillicrap, Harley,
  Silver, and Kavukcuoglu]{mnih2016asynchronous}
Volodymyr Mnih, Adria~Puigdomenech Badia, Mehdi Mirza, Alex Graves, Timothy
  Lillicrap, Tim Harley, David Silver, and Koray Kavukcuoglu.
\newblock Asynchronous methods for deep reinforcement learning.
\newblock In \emph{International conference on machine learning}, pages
  1928--1937. PMLR, 2016.

\bibitem[Corral-Acero et~al.(2020)Corral-Acero, Margara, Marciniak, Rodero,
  Loncaric, Feng, Gilbert, Fernandes, Bukhari, Wajdan, Martinez, Santos,
  Shamohammdi, Luo, Westphal, Leeson, DiAchille, Gurev, Mayr, Geris,
  Pathmanathan, Morrison, Cornelussen, Prinzen, Delhaas, Doltra, Sitges,
  Vigmond, Zacur, Grau, Rodriguez, Remme, Niederer, Mortier, McLeod, Potse,
  Pueyo, Bueno-Orovio, and Lamata]{corral2020digital}
Jorge Corral-Acero, Francesca Margara, Maciej Marciniak, Cristobal Rodero,
  Filip Loncaric, Yingjing Feng, Andrew Gilbert, Joao~F Fernandes, Hassaan~A
  Bukhari, Ali Wajdan, Manuel~Villegas Martinez, Mariana~Sousa Santos, Mehrdad
  Shamohammdi, Hongxing Luo, Philip Westphal, Paul Leeson, P.aolo DiAchille,
  Viatcheslav Gurev, Manuel Mayr, Liesbet Geris, Pras Pathmanathan, Tina
  Morrison, Richard Cornelussen, Frits Prinzen, Tammo Delhaas, Ada Doltra,
  Marta Sitges, Edward~J Vigmond, Ernesto Zacur, Vicente Grau, Blanca
  Rodriguez, Espen~W Remme, Steven Niederer, Peter Mortier, Kristin McLeod,
  Mark Potse, Esther Pueyo, Alfonso Bueno-Orovio, and Pablo Lamata.
\newblock The ‘\uppercase{D}igital twin’to enable the vision of precision
  cardiology.
\newblock \emph{European heart journal}, 41\penalty0 (48):\penalty0 4556--4564,
  2020.

\bibitem[Sharei et~al.(2018)Sharei, Alderliesten, van Den~Dobbelsteen, and
  Dankelman]{sharei2018navigation}
Hoda Sharei, Tanja Alderliesten, John~J van Den~Dobbelsteen, and Jenny
  Dankelman.
\newblock Navigation of guidewires and catheters in the body during
  intervention procedures: A review of computer-based models.
\newblock \emph{Journal of Medical Imaging}, 5\penalty0 (1):\penalty0 010902,
  2018.

\bibitem[Stepniak et~al.(2020)Stepniak, Ursani, Paul, and
  Naguib]{stepniak2020novel}
Karolina Stepniak, Ali Ursani, Narinder Paul, and Hani Naguib.
\newblock Novel 3\uppercase{D} printing technology for \uppercase{CT} phantom
  coronary arteries with high geometrical accuracy for biomedical imaging
  applications.
\newblock \emph{Bioprinting}, 18:\penalty0 e00074, 2020.

\bibitem[Vukicevic et~al.(2017)Vukicevic, Mosadegh, Min, and
  Little]{vukicevic2017cardiac}
Marija Vukicevic, Bobak Mosadegh, James~K Min, and Stephen~H Little.
\newblock Cardiac 3\uppercase{D} printing and its future directions.
\newblock \emph{JACC: Cardiovascular Imaging}, 10\penalty0 (2):\penalty0
  171--184, 2017.

\end{thebibliography}

\section*{Acknowledgement}
\label{sec:acknowledge}
This research is based upon work supported by the Ministry of Trade, Industry $\&$ Energy (MOTIE, Korea), Ministry of Science $\&$ ICT (MSIT, Korea), and Ministry of Health $\&$ Welfare (MOHW, Korea) under Technology Development Program for AI-Bio-Robot-Medicine Convergence (20001638), and by the Korea Medical Device Development Fund grant funded by Korea government (the Ministry of Science $\&$ ICT, the Ministry of Trade, Industry $\&$ Energy, the Ministry of Health $\&$ Welfare, the Ministry of Food and Drug Safety) (Project Number: KMDF$\_$PR$\_$20200901$\_$0013).

\newpage
\section*{Appendix}
\label{sec:append}

\setcounter{figure}{0}
\renewcommand{\thefigure}{A\arabic{figure}}

\begin{figure}[hbp]
	\centering
    \includegraphics[width=\textwidth,keepaspectratio]{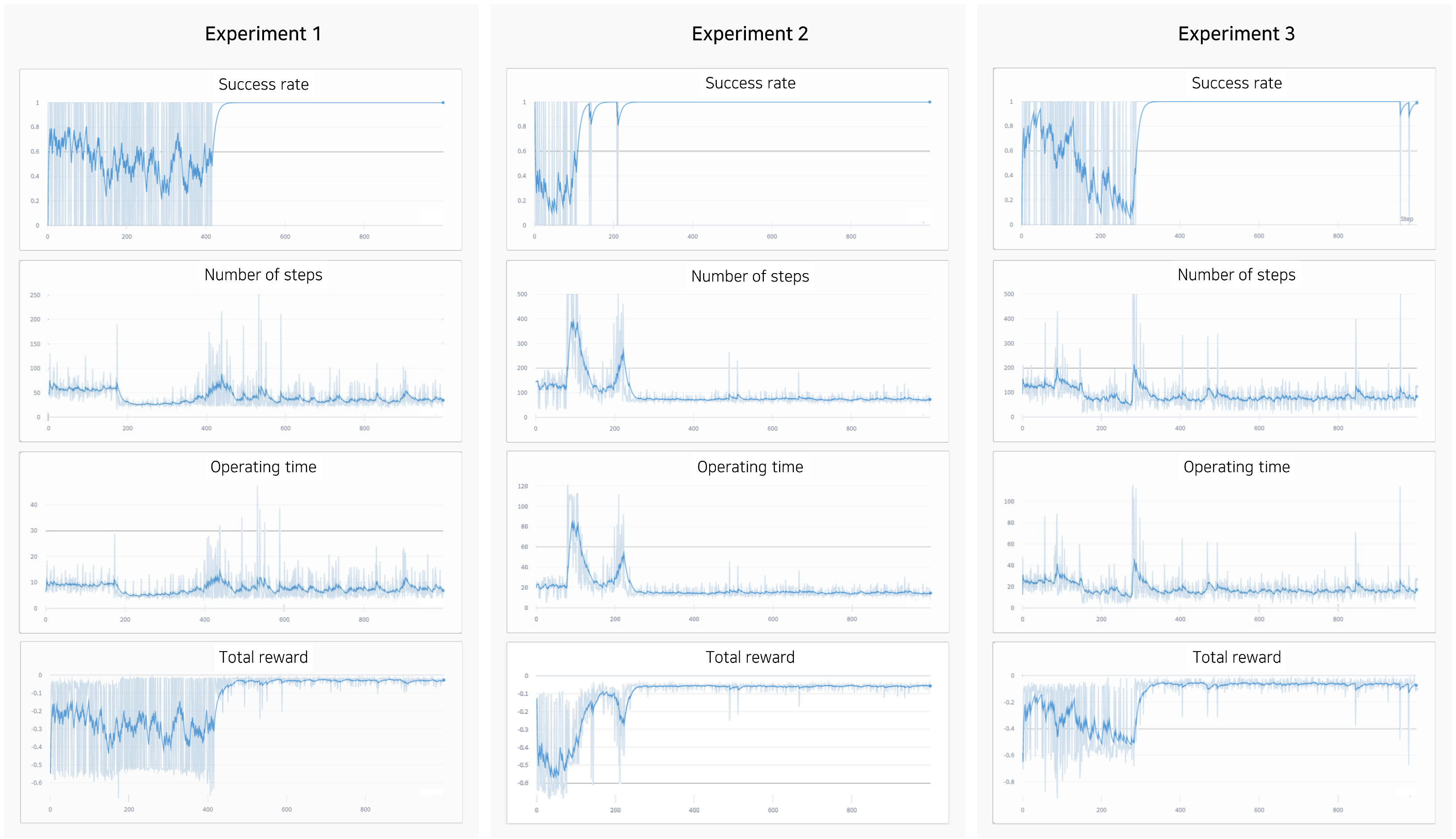}
	\caption{Success rate, number of steps, operating time, and total reward in subsequent 3 experiments using 3D phantom.}
	\label{fig:figA}
\end{figure}

\setcounter{table}{0}
\renewcommand{\thetable}{A\arabic{table}}

\begin{table}[hbp]
	\caption{Hyper-parameters for reinforcement learning agent.}
	\centering
	\begin{tabular}{lc}
		\toprule
		{Hyper-parameter}&{Value}\\
        \midrule		
		{Batch size}&{32}\\
		{Buffer size}&{100,000}\\
		{Discount factor (gamma)}&{0.99}\\
		{Update frequency}&{4}\\
		{Target network soft update ratio}&{0.005}\\
		{Gradient clip}&{10}\\
		{Multi-step returns}&{v}\\
		{Prioritization exponent alpha}&{0.4}\\
		{Prioritization important sampling beta}&{0.6}\\
		{Distributional atom}&{51}\\
		{Distributional min/max values}&{Value}\\
		{Hyper-parameter}&{[-2.0, 0.0]}\\
		{Noisy Net sigma (std)}&{0.5}\\
		{Human demo pre-training step}&{1,000}\\
		{Human demo supervised loss margin}&{0.8}\\
		{Learning rate}&{0.0001}\\
		{Weight decay}&{0.00001}\\		
		{Optimizer}&{Adam}\\	
	
		\bottomrule
	\end{tabular}
	\label{tab:atable1}
\end{table}

\end{document}